\documentclass[preprint,12pt]{elsarticle}

\usepackage{lineno,hyperref}
\usepackage{subfigure}
\usepackage{epsfig}

\usepackage{amssymb}
\usepackage{graphicx}
\usepackage{algorithm}
\usepackage{amsmath}
\usepackage{multirow}
\usepackage{diagbox}
\usepackage{colortbl}
\usepackage{booktabs}
\newcommand{\RNum}[1]{\uppercase\expandafter{\romannumeral #1\relax}}
\usepackage{textcomp,booktabs}
\journal{}

\begin{document}

\begin{frontmatter}



\title{Fusing Motion Patterns and Key Visual Information for Semantic Event Recognition in Basketball Videos}


\author[mymainaddress,mysecondaryaddress]{Lifang Wu}

\author[mymainaddress]{Zhou Yang}

\author[mymainaddress]{Qi Wang}

\author[mymainaddress,mysecondaryaddress]{Meng Jian\corref{mycorrespondingauthor}}
\cortext[mycorrespondingauthor]{Corresponding author}
\ead{jianmeng648@163.com}

\author[mymainaddress]{Boxuan Zhao}

\author[mythirdaddress]{Junchi Yan}

\author[myfourthaddress,myfifthaddress]{Chang Wen Chen}

\address[mymainaddress]{Beijing University of Technology, Beijing, China}
\address[mysecondaryaddress]{Beijing Municipal Key Lab of Computation Intelligence and Intelligent Systems, Beijing, China}
\address[mythirdaddress]{Shanghai Jiao Tong University, Shanghai, China}
\address[myfourthaddress]{School of Science and Engineering, Chinese University of Hong Kong, Shenzhen, China}
\address[myfifthaddress]{Department of Computer Science and Engineering, SUNY Buffalo, NY, USA}

\begin{abstract}
   Many semantic events in team sport activities e.g. basketball often involve both group activities and the outcome (score or not). Motion patterns can be an effective means to identify different activities. Global and local motions have their respective emphasis on different activities, which are difficult to capture from the optical flow due to the mixture of global and local motions. Hence it calls for a more effective way to separate the global and local motions. When it comes to the specific case for basketball game analysis, the successful score for each round can be reliably detected by the appearance variation around the basket. Based on the observations, we propose a scheme to fuse global and local motion patterns (MPs) and key visual information (KVI) for semantic event recognition in basketball videos. Firstly, an algorithm is proposed to estimate the global motions from the mixed motions based on the intrinsic property of camera adjustments. And the local motions could be obtained from the mixed and global motions. Secondly, a two-stream 3D CNN framework is utilized for group activity recognition over the separated global and local motion patterns. Thirdly, the basket is detected and its appearance features are extracted through a CNN structure. The features are utilized to predict the success or failure. Finally, the group activity recognition and success/failure prediction results are integrated using the kronecker product for event recognition. Experiments on NCAA dataset demonstrate that the proposed method obtains state-of-the-art performance.
\end{abstract}

%


\end{frontmatter}

\section{Introduction}

Recognizing semantic events in sports videos has received increasing attention from the researchers of computer vision due to its great challenges and wide potential applications in real world. The main challenge is how to extract discriminative and robust spatio-temporal contextual features in the dynamic scenes. A great deal of attempts utilized various modalities of data to establish the mapping relationship between the group activity and semantic representation such as key components \cite{1}, multi-level interaction representations \cite{2}, hierarchical relational representations \cite{3} and semantic graph \cite{4}.

In basketball videos, representations of visual appearance tend to be unreliable due to variation of play ground, player's suits and over-complicated background. Challenges such as occlusion between individuals and motion blur problems greatly restrict the performance of player detection and tracking based approaches \cite{1, 2, 3, 4}. In this circumstance, it is more promising to model representations on the dynamic information to avoid the bias of appearance variance in the chaotic scene and focus only on the overall motion regularities.

\begin{figure}
\begin{center}
   \includegraphics[width=0.7\linewidth]{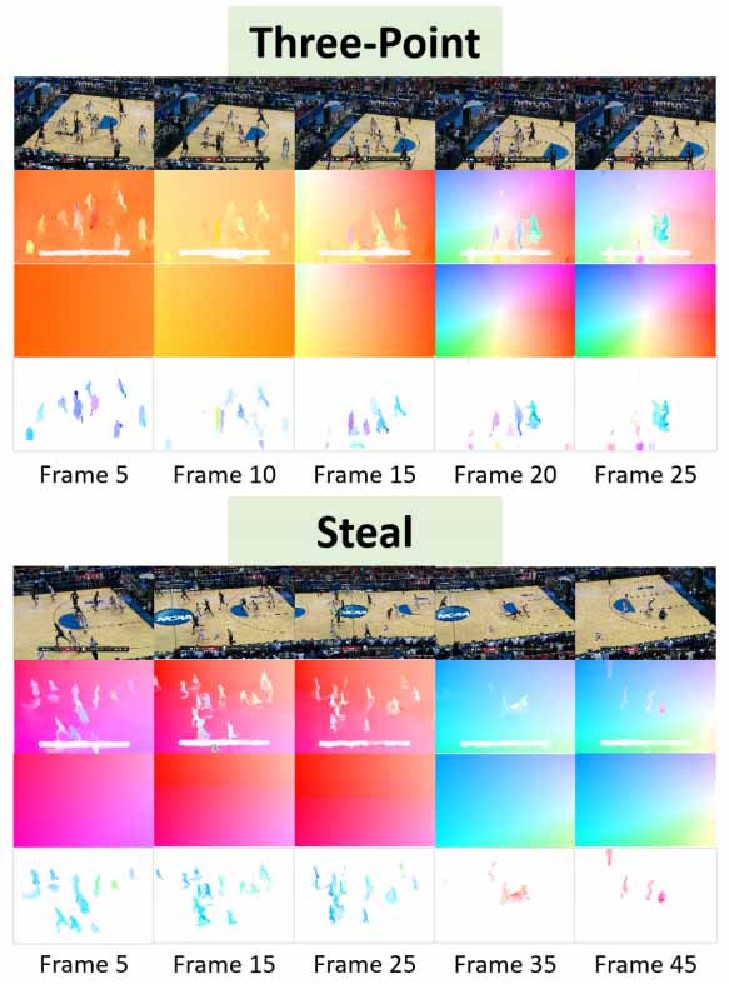}
\end{center}
   \caption{Visualization results of frames and motion patterns of 3-point and steal activities for basketball videos in NCAA dataset \cite{1}. In both cases, from top to bottom each row denotes video frames, mixed motion, separated global and separated local motions respectively. For better visualization, We use the color-coding rule in \cite{48} to encode motions where different color represents different motion directions while the saturation represents the motion amplitude. In 3-point activity, camera is firstly moving in translation, followed by zooming in while the players gradually concentrate to the basket for rebound. As for steal events, both global and local motion directions will be reversed at some point.}
\end{figure}

In basketball games, the camera mostly focuses on the game by `translate movement' or `zooming in/out'. For a specific group activity, the camera is adjusted in almost the same way. It conveys the consistency of global motion patterns in the basketball videos. As a typical kind of team sport, a group activity can be represented as an offensive-defensive confrontation. And the tactics in the basketball games are reflected as players' position distributions. To some extent, the tactics could be represented as local motion patterns. The global and local motion patterns of two sample activities are shown in Figure 1. In three-point activity, the general camera motion pattern can be summarized as pan or tilt (corresponding to the translation in the videos) at the beginning, then followed by zooming in (corresponding to the centrifugal motion in the videos). In the meantime, the position distribution of the players is firstly scattered and then concentrated in the direction of the basket for rebound. In steal activity, the overall local motion patterns of the players show a sudden reverse. The camera follows this change and reverses its translation direction simultaneously. A specific kind of group activity is much related to the global or local motions. However, the motions from the estimated optical flow are the mixture of global and local motions. It is necessary to separate them from the mixed motions. Global motion estimation \cite{57, 58, 59} is a typical issue in computer vision. In this paper, we orient a novel global motion estimation scheme for scene of basketball games based on our observations. On the other hand, the success or failure of the activity corresponds to whether the ball falls into the basket or not, and whether the basket net moves or not. Therefore, the outcome is mainly related to the key local visual information (KVI), as shown in Figure 2.

\begin{figure}
\begin{center}
   \includegraphics[width=0.7\linewidth]{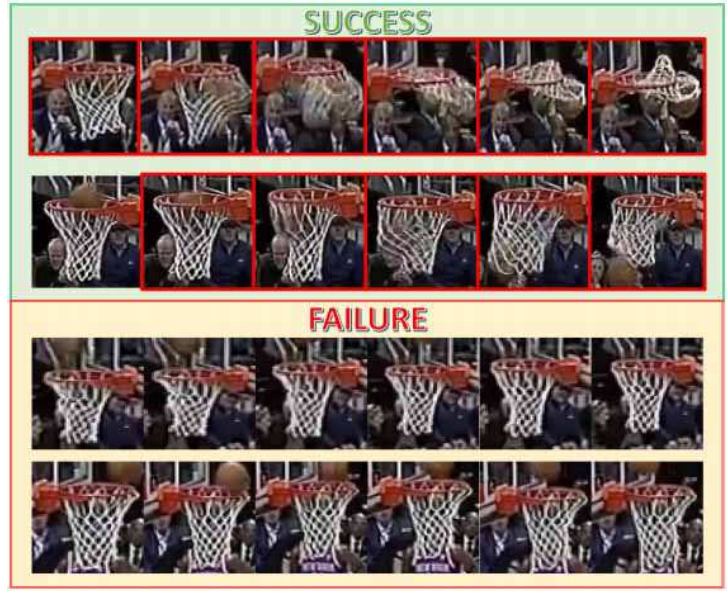}
\end{center}
   \caption{Examples of success and failure for key visual region. Each line shows the last 6 frames from success or failure clip. The pictures with red borders are frames with great visual distinctions in success clips.}
\end{figure}

Motivated by the above, we develop a unified framework to integrate the dynamic features of the global and local motion patterns (MPs) with the appearance features from key visual information (KVI) as illustrated in Figure 3. Specifically, the video frames are fed into two pipelines. The first one involves separating the global and local motions, and extracts the dynamic motion pattern representation with a two-stream 3D CNN model for group activity recognition. In the other pipeline, KVI is extracted using CNN model from the detected basket regions for success/failure prediction. The group activity recognition and success/failure prediction from two pipelines are further integrated (by the Kronecker product operation) to infer the semantic events in basketball videos.

\begin{figure}
\begin{center}
   \includegraphics[width=1.0\linewidth]{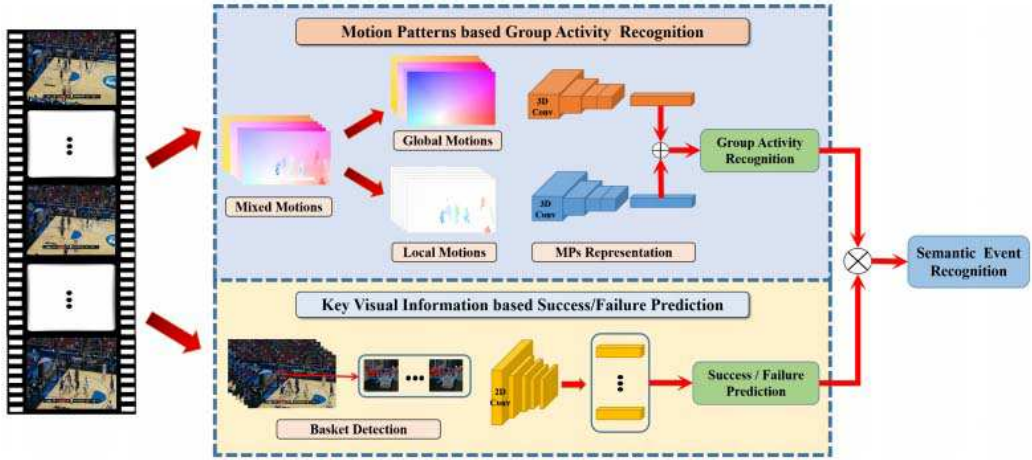}
\end{center}
   \caption{An overview of our proposed fusion framework for separated motion patterns and key visual information.
   Video frames are input into two pipelines. In the first pipeline, global and local motions are firstly separated from the mixed motions. Then, motion patterns are represented through a two-stream 3D CNN module for group activity recognition. In the second pipeline, key visual information is extracted by CNN with the input of detected basket regions. Finally, prediction vectors from two pipelines are integrated for semantic events recognition in basketball videos. Here Kronecker products are used without loss of generality.}
\end{figure}

In a nut shell, the main contributions of this work can be summarized as follows:

1) We make a basic observation that the motion in the basketball game (and also many other games) contains both global motion and local movement. Furthermore, the global motion is caused by the camera movement, which can reflect the intention of the camera man and vary depending on the type of matches. While the local ones show the tactics of the team. Thus the local motion or the tactics of the same activity are similar. Based on the observations, we devise a simple yet effective method to estimate the camera motion, such that the global and local motions are effectively separated from the mixed motions.

2) We present a unified framework to fuse global and local motion patterns (MPs), as well as key visual information (KVI) for semantic basketball video event recognition. The KVI module is tailored to detect the basket appearance change which indicates the outcome for a basketball game round.

3) Experimental results on the public NCAA dataset~\cite{1} demonstrate that the proposed approach outperforms the state-of-the-arts due to its advantages in employing MPs and KVI via ablation study.

The paper is organized as follows. Section \ref{sec:related} discusses the related work in literature. Section \ref{sec:method} presents the proposed method which is based on our observation on the motion patterns and the local appearance for the score outcome. Experiments are conducted in Section \ref{sec:exp} and Section \ref{sec:con} concludes this paper.

\section{Related Work}\label{sec:related}

\textbf{Group/Collective Activity Recognition:}
Before the advent of powerful deep learning methods, devising effective hand-crafted features for group activity recognition has been extensively studied, which commonly represent the sequence images with a number of predefined descriptors \cite{6,7,8,9,10,11,12,56}. Lan et al. \cite{6} proposed a latent variable framework to jointly model the contextual information of group-person and person-person interaction. Shu et al. \cite{12} conducted a spatiotemporal AND-OR graph to jointly infer groups, events and individual roles. Wang et al. \cite{7} proposed a bi-linear program based MAP inference with a latent structural SVM for human activity modelling.

More recently, deep neural network architecture achieves substantial success on group activity understanding due to its high-capacity of multi-level feature representation and integration \cite{1, 2, 3, 4, 14, 15}. Wang et al. \cite{2} proposed a recurrent interactional context model to aggregate person level, group level and scene level interactions. Shu et al. \cite{15} introduced a confidence-energy recurrent network (CERN) which utilized a p-value based energy regularization layer to estimate the minimum of the confident energy. Deng et al \cite{14} performed structure learning by a unified framework of integrating graphical models and a sequential inference network.

\textbf{Sport Video Analysis:}
Recently, a considerable amount of efforts have been devoted to team sports analysis, such as basketball \cite{1}, volleyball \cite{3, 4, 5, 15, 16, 17, 53, 54, 55}, soccer \cite{21, 22}, water polo \cite{20}, ice hockey \cite{23} etc. Ramanathan et al. \cite{1} introduced an attention based BLSTM network to identify the most relevant component (key player) of the corresponding event and recognize basketball events. Ibrahim et al. \cite{3} built a relationship graph guided network to infer relational representations of each player and encode scene level representations. Wu et al. \cite{47} proposed a domain knowledge based basketball semantic events classification method which represented global and collective motion patterns on different event stages for event recognition.

\textbf{Global and local motion separation:}
In broadcast sports videos, global motion tends to be dominant. Therefore, it is intuitively motivated to firstly separate global motion and local motion from the mixed motion fields. In the previous works, \cite{25, 26} extracted dense trajectories in the dense flow field to eliminate the camera motion filed. However, they employed hand-craft feature descriptors to match feature points which can be time-consuming and less applicable to deploy on the real time applications. In \cite{28}, Hasan et al. proposed a nonparametric camera descriptor which adopted data statistics to localize the local motion region to characterize global motion representation. In \cite{27}, Yu et al. assumed that camera motion consisted of only simple translate movement and subtracted the estimated motion field from the flow field to obtain local motion field. However, in broadcast sports videos, camera motions are usually the combinations of multiple basic shot motion patterns.

In summary, most existing group activity recognition methods depend on both the action detection \cite{16, 18} or tracking \cite{1} feeding to appearance stream, as well as the local motion field feeding to motion stream, to extract either multi-level objects features or contextual spatiotemporal interactional representations. Nevertheless, under certain circumstances, feature representation suffers a lot from frequent occlusion among players, rich variations of player poses and camera movement. These challenges make the high-order contextual information modelling even difficult. We address the problem by devising a way of joint representation of motion patterns and key visual information for both group activity and semantic event recognition in basketball videos.

\section{The Proposed Basketball Video Event Recognition Approach}\label{sec:method}
\subsection{Observations}
We delve into the characteristics in basketball videos and make the flowing observations:

\textbf{1) Global motion in the basketball video mostly refers to the camera movement which reflects the ways that the cameramen represent the group activities.} The camera movement is frequent in basketball videos. Specifically, the camera will keep focusing on the hot area of the court by lens adjustment. Generally, different activities have unique focusing areas and motion severities on the court. As shown in Figure 1, in three-point activity, the camera is firstly translated to track the player who possesses the basketball, and the camera will zoom in around the basket to obtain a clear view for audience when the shooting event is performed. Also, in steal event, the camera will show a reverse motion when the steal action happens. However, in different type of matches, such as NCAA, NBA or CBA, it is possible that the camera movement (global motion) shows a little distinction.

\textbf{2) Local motion in the basketball video mostly refers to the team tactics and the collective motion of players.}
Basketball game is a type of tactical team-sport in which the local motions present the specific tactics. When the offensive players organize an attack, the defence players will show corresponding movement. For example, in three-point activity, defenders may take one-on-one defence and the position distribution is relatively scattered and the confrontation is not much intense. After the shot, all the players will struggle for position and rebound which may cause intense motions. In steal event, when one of the defenders steals the ball, he/she will quickly launch a counterattack. Meanwhile, all the other players will switch the role and rush to the other side of the court.

\textbf{3) Success/failure for each round is correlated to the appearance variation in the key region around the basket.}
If the shooting is successful, the ball will fall into the basket along with the motion of the net. Failure means the ball is bounced off or not touching the basket, which generally corresponds to the static basket net. Therefore, Success/Failure could be identified from the key visual information (appearance variation in basket region).

Based on the above observations, we propose an MPs and KVI fusion scheme for semantic event recognition in basketball videos. The proposed two-pipeline structure is shown in Figure 3. In the pipeline for group activity recognition, global and local motions are input into two separated 3D CNN structures for motion pattern representation. Two modalities of motion patterns are integrated at the softmax layers of the two networks using the late fusion strategy. In the success/failure prediction pipeline, the basket detection results of input frames are fed into a CNN structure for key visual information representation. The prediction vectors from the two pipelines are further fused by the Kronecker product operation to obtain the semantic event recognition results in basketball videos.

\subsection{Global and Local Motion Separation}

Global and local motions have their own emphasis on different activities. The global motions can effectively present the activities such as three-point, steal and so on, while the local motion is more discriminative to group activity such as 2-point and layup. Therefore, it is possible to promote the representation capability to the corresponding activities using the separated global and local motions. In this section, we will firstly estimate the global motions from the mixed motions, then obtain the local motion from the mixed and the global motions.

\begin{figure}
\centering
\subfigure[]{
\label{Fig.sub.1}
\includegraphics[height = 1.0in,width = 1.0in]{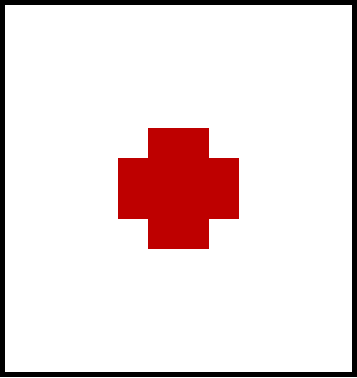}}
\subfigure[]{
\label{Fig.sub.2}
\includegraphics[height = 1.0in,width = 1.0in]{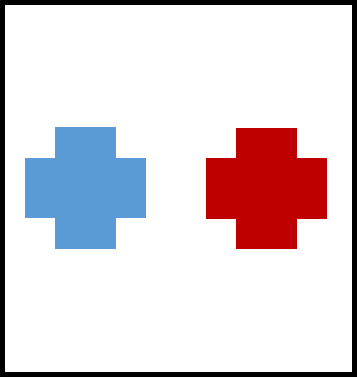}}
\subfigure[]{
\label{Fig.sub.3}
\includegraphics[height = 1.0in,width = 1.0in]{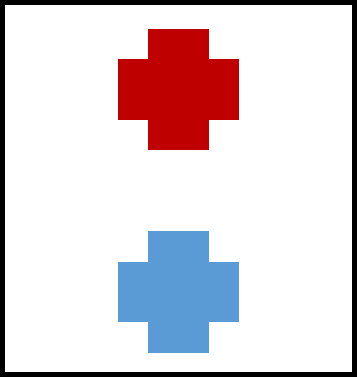}}
\subfigure[]{
\label{Fig.sub.4}
\includegraphics[height = 1.0in,width = 1.0in]{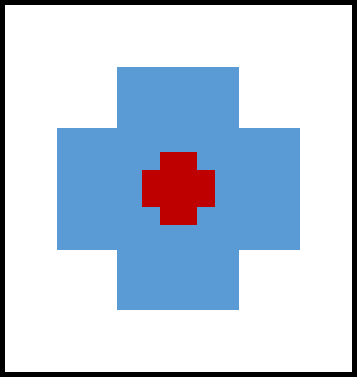}}
\caption{Basic camera motions. We use the crosses to simulate the change of the actual object under different camera motions. (a) Static. (b) Pan right / left. (c) Tilt up / down. (d) Zoom in / out.}
\label{Fig.lable}
\end{figure}

\subsubsection{Global Motion Estimation}
The camera adjustment is the main contributor to the global motions. Basic camera adjustments include static, tilt (up or down), pan (left or right) and zoom (in or out) as shown in Figure 4. The basic camera adjustments corresponds to the global motions which can be represented as the corresponding color coding images, as shown in Figure 5. We use color-coding images to represent the camera motions that different color level represents different motion directions while the color intensity or saturation represents the motion amplitude.

\begin{figure}
\centering
\subfigure[]{
\label{Fig.sub.1}
\includegraphics[height = 0.8in,width = 0.8in]{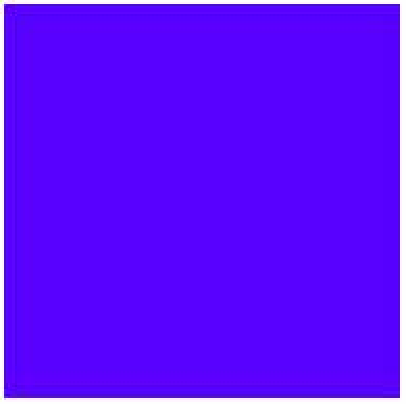}}
\subfigure[]{
\label{Fig.sub.2}
\includegraphics[height = 0.8in,width = 0.8in]{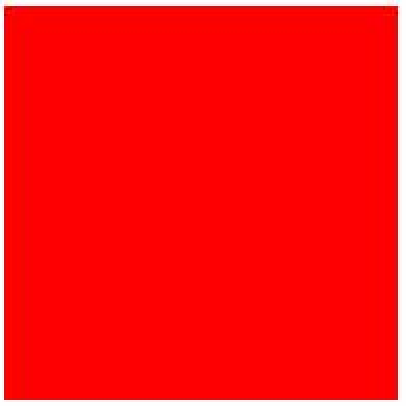}}
\subfigure[]{
\label{Fig.sub.3}
\includegraphics[height = 0.8in,width = 0.8in]{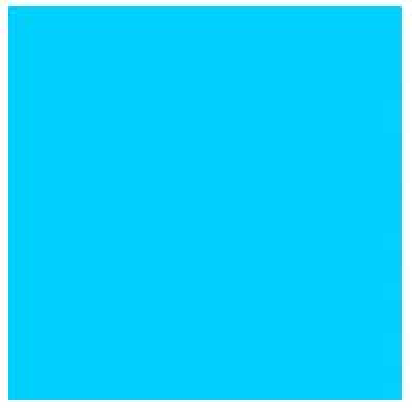}}
\subfigure[]{
\label{Fig.sub.4}
\includegraphics[height = 0.8in,width = 0.8in]{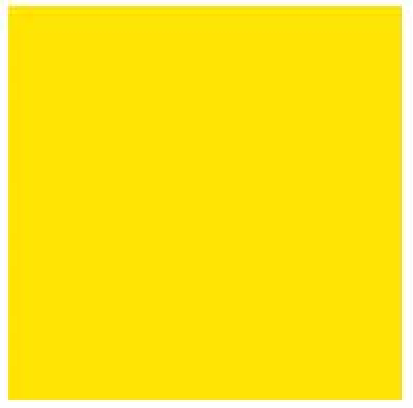}}
\subfigure[]{
\label{Fig.sub.5}
\includegraphics[height = 0.8in,width = 0.8in]{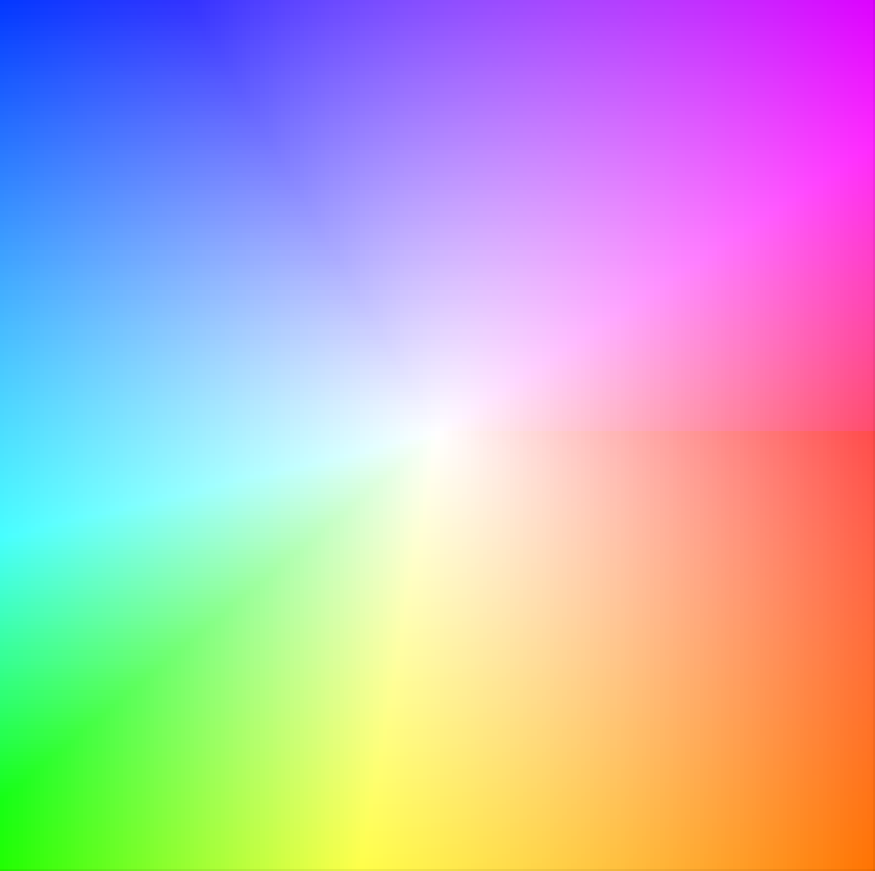}}
\subfigure[]{
\label{Fig.sub.6}
\includegraphics[height = 0.8in,width = 0.8in]{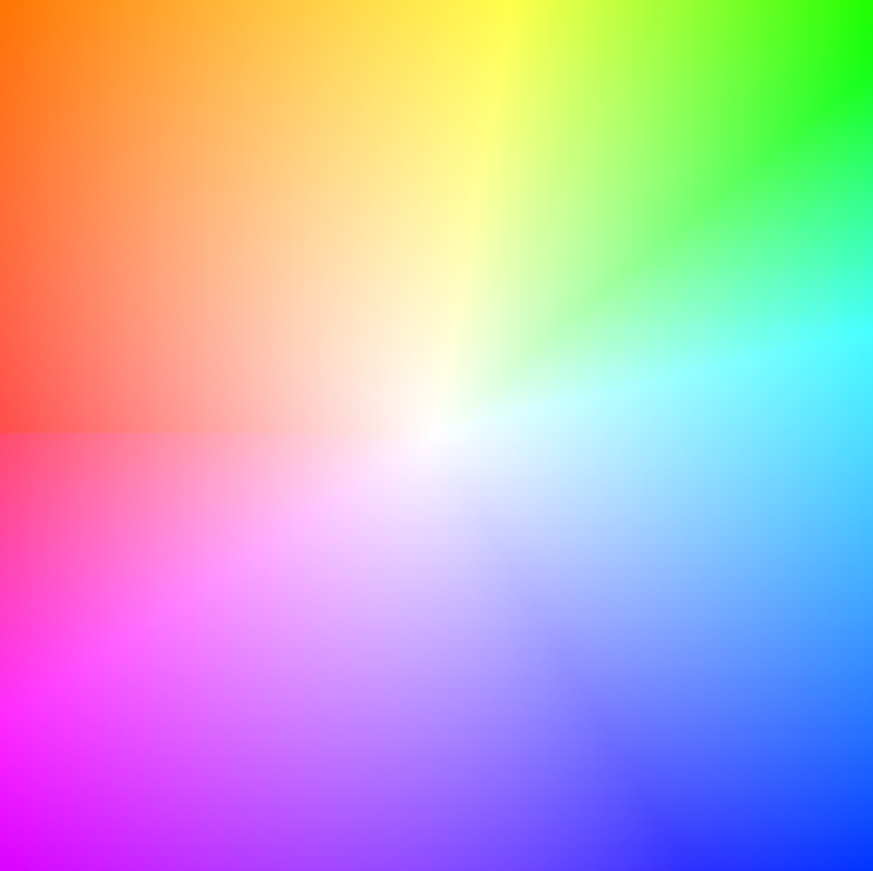}}
\caption{Basic camera motions in the format of color coding: (a) Pan left. The global motion appears as right movement with the same amplitudes in all pixels. (b) Pan right. The global motion appears as left movement with the same amplitudes in all pixels. (c) Tilt up. The global motion appears as upward movement with the same amplitudes in all pixels. (d) Tilt down. The global motion appears as downward movement with the same amplitudes in all pixels. (e) Zoom in. The global motion appears as the movement toward the center point in radial direction. The motion amplitude decreases from the center to the perimeter (f) Zoom out. The global motion appears as the movement away from the center point in radial direction. The motion amplitude decreases from the center to the perimeter.}
\label{Fig.lable}
\end{figure}

\begin{figure}
\centering
\subfigure[]{
\label{Fig.sub.1}
\includegraphics[height = 1.7in,width = 2.5in]{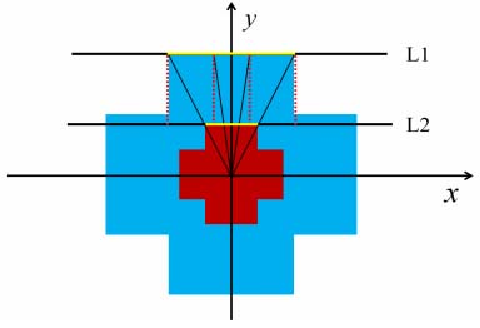}}
\subfigure[]{
\label{Fig.sub.2}
\includegraphics[height = 1.7in,width = 2.5in]{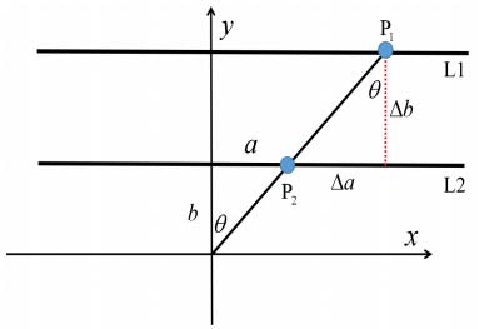}}
\caption{(a) The illustration of the displacement of the corresponding points in the scene after zooming. The red cross represents the original object while the blue cross represents the object after zooming in transformation. (b) A detailed geometric relationship diagram of zooming in transformation.}
\label{Fig.lable}
\end{figure}


Then, we deeply analyze the intrinsic property of the global motion under different camera transformations by summarizing the displacement regularity of the corresponding points in the motion fields. When the camera moves as pan or tilt, all the points in the motion fields have the same motion amplitude in both X and Y coordinate. For zooming transformation, as shown in Figure 6(a), the global motion appears as the movement toward the center point in radial direction. $L1$ and $L2$ are two parallel lines. The yellow lines on $L2$ and $L1$ at the top of the red and blue cross represent the mapping of the corresponding points after zooming in transformation. In this case, we further deduce the formula of point displacement in both X and Y direction as illustrated in Figure 6(b). Point $P_2(x_0, y_0)$ on line $L2$ represents the point in the red cross in Figure 6(a) while point $P_1(x_0+\Delta x_0, y_0+\Delta y_0)$ on line $L1$ represents the corresponding point after zooming in transformation. It is obvious that the points on line $L2$ have the same Y direction displacement after zooming. For displacement along X direction, it can be formulated through geometry deduction as equation (1) and equation (2).

\begin{equation}
\tanh(\theta)= \frac{a}{b} = \frac{\Delta a} {\Delta b}
\end{equation}

\begin{equation}
\Delta a = \frac{\Delta b}{b}a
\end{equation}

Thus, the displacement of the points along X direction is a linear transformation. Considering a point in current frame (anchor frame) with the coordinate $(x, y)$, mapping to $(x', y')$ in the reference frame (target frame) after global motion transformation. The global motion model can be written as:

\begin{equation}
\begin{aligned}
x' = m_0x+m_1 \\
y' = m_2y+m_3
\end{aligned}
\end{equation}

where $[m_0, m_1, m_2, m_3]$ are model parameters. Then, the motion vector of point $P$ can be denoted as $MVp=(m_{0}x_0+m_{1}, m_{2}y_0+m_{3})$. In the actual scenes, the camera movements can be seen as the combinations of the basic camera movements, that is, linear superposition of the basic camera transformations. Therefore, the motion amplitude of any points in the global motion field can be computed through the above transformations. Furthermore, we discuss the amplitude distribution of the x- and y-component of the global motion field. The motion field records the motion increments of each pixel which can be represented as:
\begin{equation}
\begin{aligned}
\Delta x = x'-x = (m_0-1)x+m_1 \\
\Delta y = y'-y = (m_2-1)y+m_3
\end{aligned}
\end{equation}

In conclusion, we draw the following characteristics of the global motion field: (1) In x-component of global motion field, the points with the same X coordinate have the same amplitude while the points with the same Y coordinate are subject to linear variation. (2) Similarly, in y-component of global motion field, the points with the same Y coordinate have the same amplitude while the points with the same X coordinate are subject to linear variation. (3) In basketball videos, local motions results from the movement of the athletes. They only occupy a small region in the relatively center of the scene. Moreover, the mixed motion in the marginal region of the image is global motion with high probability.


Theoretically, based on the above characteristics, if the motion vectors at four corners of the motion field are obtained, global motions of the whole image could be estimated through linear interpolations. Therefore, we propose a scheme for global motion estimation with the input of mixed motion field. Given a mixed motion field with the shape of $H\times W\times 2$, the four corner points can be presented as $P_{11}(x_{11}$, $y_{11})$, $P_{1W}(x_{1W}$, $y_{1W})$, $P_{H1}(x_{H1}, y_{H1})$ and $P_{HW}(x_{HW}, y_{HW})$ (the index starts from 1). Firstly, motion vectors at the four corners of the mixed motion field could be computed by statistics. Given a sequence of x-component in the column of left edge in the mixed motion field, we sort all the data in the sequence and assign the mean value of the data in the middle 60\% of the sequence to x-component in this column. Following the same operation, we could obtain the x and y-components of the four corners in global motion field. Secondly, the x or y-components of the points in the four marginal lines in global motion field are further obtained by interpolation from the corresponding end points. Thirdly, we utilize the linear interpolation algorithm to estimate the x and y-components of global motions for all of non-marginal points in the motion field. The results are shown in the third row in Figure 1.


\subsubsection{Local Motion Estimation}

\begin{figure}
\centering
\subfigure[]{
\label{Fig.sub.1}
\includegraphics[height = 1.0in,width = 1.5in]{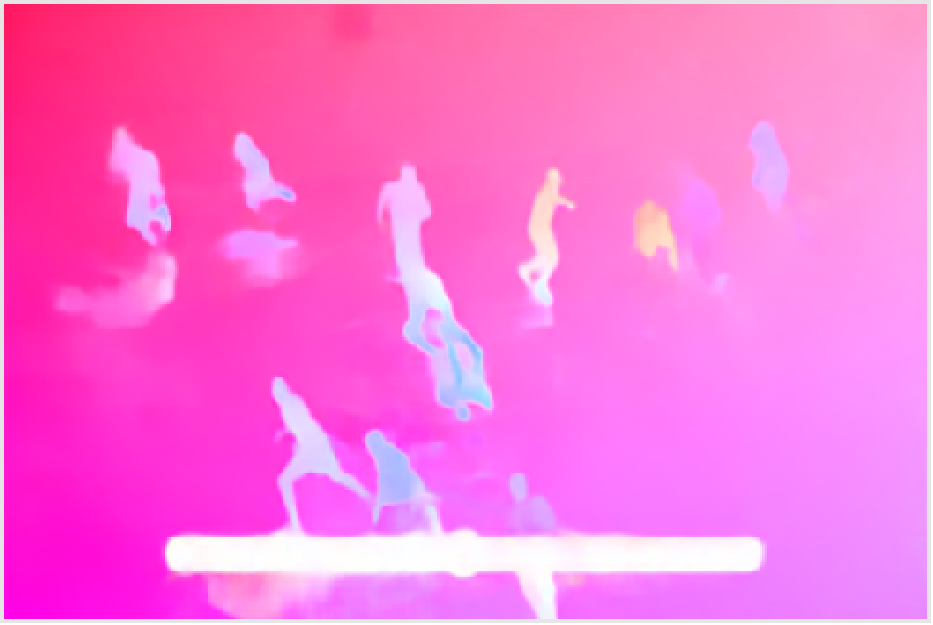}}
\subfigure[]{
\label{Fig.sub.2}
\includegraphics[height = 1.0in,width = 1.5in]{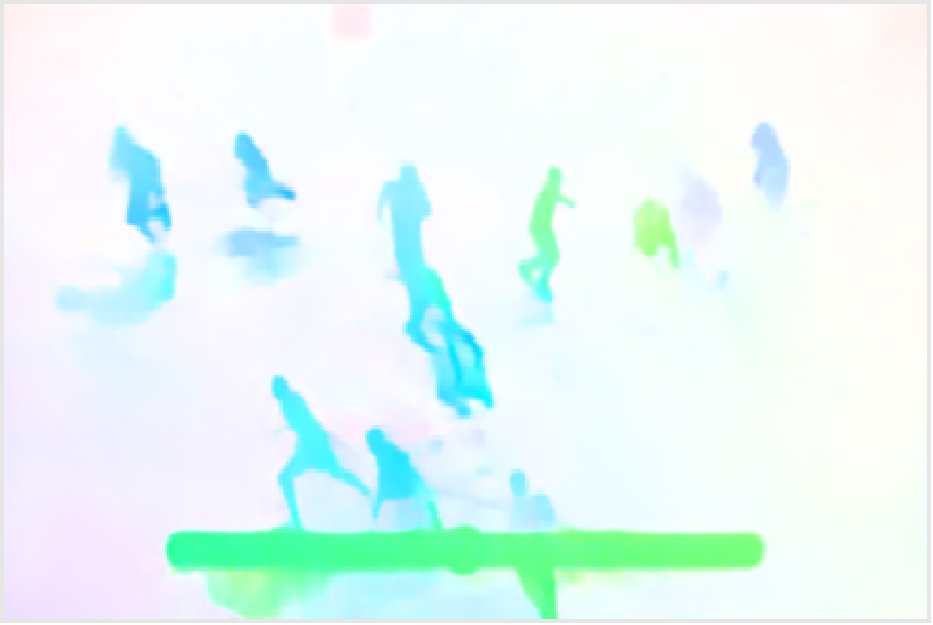}}
\subfigure[]{
\label{Fig.sub.3}
\includegraphics[height = 1.0in,width = 1.5in]{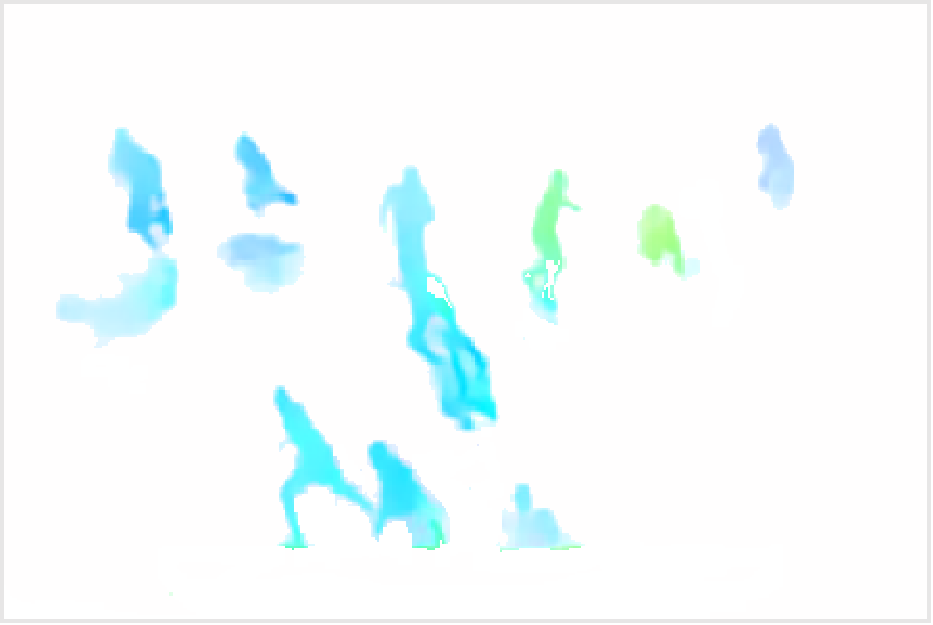}}
\caption{Local motion estimation result: (a) Mixed motion. (b) Local motion estimated by directly subtracting the global motion from the mixed motion. (c) Threshold based local motion estimation.}
\label{Fig.lable}
\end{figure}

It seems a natural way to disentangle the local motion from the mixed flow filed by subtracting the global motion field. However, this simple operation may produce some extra noises like the still scoreboard regions in the scene or minor moving amplitude. To alleviate this problem, we devise a threshold based scheme for local motion estimation.

Assuming $(x_{m}^i, y_{m}^j)$, $(x_{g}^i, y_{g}^j)$ and $(x_{l}^i, y_{l}^j)$ as the points in mixed flow field, global motion field and local motion field respectively with the coordinate $(i, j)$. And $r^{(i,j)}=\sqrt{{x^i}^{2} + {y^j}^2}$ is the resultant motion amplitude of point $(i, j)$. The local motion field can be estimated as follows:

{\setlength\abovedisplayskip{1pt}
\setlength\belowdisplayskip{1pt}
\begin{equation}
\begin{aligned}
x_{l}^i = \left\{ \begin{array}{lll}
 0 & |r_{m}^{(i,j)}| \leq \theta \\
 x_{m}^i - x_{g}^i & |r_{m}^{(i,j)} - r_{g}^{(i,j)}| > \theta \\
 0 & |r_{m}^{(i,j)} - r_{g}^{(i,j)}| \leq \theta \\
  \end{array} \right.
\\
y_{l}^j = \left\{ \begin{array}{lll}
 0 & |r_{m}^{(i,j)}| \leq \theta \\
 y_{m}^j - y_{g}^j & |r_{m}^{(i,j)} - r_{g}^{(i,j)}| > \theta \\
 0 & |r_{m}^{(i,j)} - r_{g}^{(i,j)}| \leq \theta \\
  \end{array} \right.
\end{aligned}
\end{equation}
}

where $\theta$ represents the threshold and are set to 1.0 in practice which means that we ignore the points moves less than 1 pixel. The local motion estimation results are shown in the forth row in Figure 1. To verify the effectiveness of threshold based local motion separation scheme, we conduct an experiment to compare the result between the local motion estimated by directly subtracting the global motion from the mixed motion and local motion computed by our method. As shown in Figure 7, by applying thresholds, the noise motion component and the irrelevant scoreboard region could be successfully suppressed.

To demonstrate the generality of our approach, we show global and local motion separation results on examples in UCF-101 dataset \cite{46} as shown in Figure 8. Notably, the global motion and local motion is isolated from the mixed motion which demonstrates that our scheme can be applies to common videos without any limitation.

\begin{figure}
\begin{center}
   \includegraphics[width=0.8\linewidth]{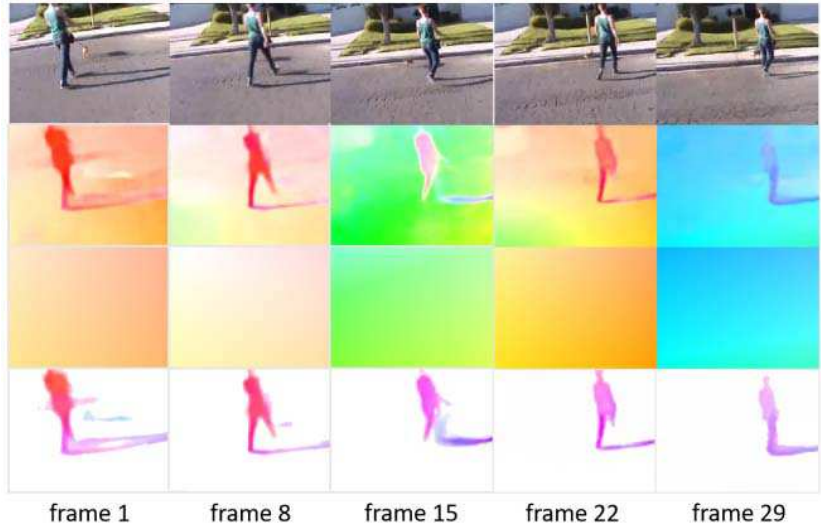}
\end{center}
   \caption{Visualization results of the global and local motion separation results on videos from UCF-101 dataset. The first row is the video frames. The second row is estimated mixed flow field. The third and the fourth rows are separated global and separated local motions.}
\end{figure}

\subsection{Fusing MPs and KVI for Event Recognition}
\subsubsection{Group Activity Recognition via Motion Patterns}


Recently, two-stream 3D CNN \cite{32, 33} structures \cite{35, 36} and multi-stream 3D CNN framework \cite{31, 34, 37, 38} have been a popular building block for sequential data modeling. Inspired by these previous works, we leverage two-stream 3D CNN structure to integrate the global and local motion patterns for group activity recognition. The network architectures in both streams are the same. Each of the 3D CNN structure has five 3D convolutional layers with 64, 128, 128, 256, 256 feature maps, followed by three fully connected layers of dimensions 2048, 2048 and number of classes. Batch normalization layer \cite{44} is joined after every convolution layers and first two fully connected layers. We use max pooling in all pooling layers with $2\times 2\times 2$ kernel size except for the first pooling layer where the kernel size is set to $2\times2\times1$. Given a video clips, the global motion and local data are sequentially fed into their corresponding network stream. The output is the probability distribution to the six group activities (3-point, Free-throw, Layup, 2-point, Slam dunk and Steal).
\\

\subsubsection{Success/Failure Prediction using Key Visual Information}

Success/Failure are the results of group activities. Success or failure can be effectively identified by whether the ball falls into the basket along with motion of the net as shown in Figure 2. According to this point, a Key Visual Information based scheme is constructed for success/failure prediction. The basket regions are detected and the appearance features of this regions are extracted for success/failure prediction. For real-time efficiency, the Single Shot Detection (SSD) model \cite{30} is employed for basket detection and AlexNet \cite{39} structure is utilized for appearance features extraction of the basket. Moreover, considering the fact that the ball in the basket only lasts for a few frames in success, we devise a strategy that the clip is considered to be success as long as one of the frame in the clip is predicted to success, otherwise, it is regarded as failure.
\\

\subsubsection{Fusion of the Two Pipelines}
From the above two pipelines, we could obtain the group activity and success/failure prediction results. We encode these two results into binary vectors. The element of the highest probability is assigned as 1 while other elements are assigned as 0. Then, we utilize Kronecker product operation to fuse these two results into semantic events recognition results which can be presented as:

\begin{equation}
V_{event}=V_{activity} \otimes V_{sf}
\end{equation}

where $ \otimes $ represents the Kronecker products operation, $V_{activity}$ is the group activity prediction result vector, $V_{sf}$ is the success/failure prediction vector and $V_{event}$ is the fusion results for semantic events prediction in basketball videos. In NCAA dataset, there are 6 types group activities including 3-point, free-throw, 2-point, layup and steal. Each of group activity can be success or failure except for steal activity. The shape of $V_{activity}$, $V_{sf}$  and $V_{event}$ are $1\times 6$, $1\times 2$ and $1\times 12$ respectively. When reporting our final 11-class semantic events recognition results, we merge steal success and steal failure in $V_{event}$.
\\

\subsubsection{Implementation Details}
~\\
\textbf{Group Activity Recognition.}
Both 3D CNN models of the two streams are firstly pretrained on Sports-1M dataset \cite{42} and then separately fine-tuned on separated global and local motion fields. We use the Adam Gradient Descent \cite{45} on each mini-batch with an initial learning rate of 0.001 and negative log likelihood criterion. The overall objective loss function is the standard categorical cross-entropy loss.

{\setlength\abovedisplayskip{1pt}
\setlength\belowdisplayskip{1pt}

\begin{equation}
L = -\sum_{i=1}^Cy_{i}log(p_i)
\end{equation}
}

where $C$ is the number of group activity classes, $y_i$ is the label and $p_i$ is the output of the softmax layer. The fusion weights ratio of two streams in Softmax layer is set to 1: 1. On the data imbalance problem in NCAA dataset, we adopt a re-sampling scheme to ensure that the number of data from 6 basic events remains equally in every mini-batch. In practice, the batch size is set to 18 (3 samples per class) due to the limitation of GPUs. Random crop scheme is employed for data augmentation. We resize the shortest edge of the frame to 112 and then crop $112\times 112$ regions randomly as spatial resolution. The length of input clip is 16 frames and the size of input data is $112\times 112\times 3\times 16$. For optical flow estimation, there exists a number of efficient deep learning based optical flow estimation methods \cite{29, 50, 52} and we adopt off-the-shelf PWC-Net \cite{52} to make the trade-off between speed and performance.
~\\
\textbf{Success/Failure Prediction.}
We manually annotate 2000 images labeling both the basket coordinate and the success or failure property for key area detection and events property prediction tasks respectively. We utilize the SSD model pretrained on VOC dataset and AlexNet framework pretrained on ImageNet \cite{43} to finetune our data. The initial learning rate is set to 1e-4 and 1e-3 for detection and classification task respectively. Adam Gradient Descent is used for learning rate adjustment. All the deep models are implemented on Pytorch deep learning framework with Nvidia Titan X GPU.

\section{Experiments}\label{sec:exp}
In this section, we conduct experiments to verify the effectiveness of the proposed method in event recognition. First, we measure the role of different motion patterns in performance of group activity recognition. Then we evaluate the performance the KVI based event success/failure predictor. Finally, the results of our proposed MPs and KVI based scheme are compared with state-of-the-arts on NCAA benchmak \cite{1, 47} for semantic events recognition in basketball videos.

\subsection{Dataset}
\textbf{The NCAA dataset}~\cite{1} is a large-scale dataset for semantic events recognition in broadcast NCAA basketball games. It contains 257 basketball games collected from YouTube. The start and end time point of each event are well-defined and provided. Based on the characteristics of semantic events, we split 11 semantic events into 6 group activities including 3-point, free-throw, layup, 2-point, slam dunk and steal and success/failure (score or not) of the activity. Besides, steal activity does not distinguish between success and failure. We follow the same training and testing sets configuration as \cite{1} including 212 games for training, 12 games for validation and 33 games for testing.

\textbf{The NBA\&CBA dataset}~\cite{47} aims at evaluating the generalization ability of group activity recognition algorithms by cross-dataset testing. The videos are from CBA and NBA basketball games, including totally 329 clips with the same annotation as NCAA dataset. All the clips are utilized for testing.

\subsection{Group Activity Recognition}

To verify the effectiveness of motion pattern representations on basketball activity recognition, we conduct several experiments based on different modalities of motion patterns and compare the performance with GCMP method \cite{47}. We also directly use the mixed motions to train a 3D CNN model to validate the benefit of the motion separation scheme for group activity recognition. Accuracy(ACC) and mean average precision (MAP) to evaluate the method. In this paper, MAP (mean average precision) is the average precision of every class. For a certain class, precision of a class refers to the proportion of the number of correctly predicted samples in the number of all the samples that are predicted to this class. Suppose a dataset contains $k$ classes (class 0 to class k-1). For a certain class $i$, assuming that the number of samples correctly classified to this class is $S_T^i$ and the number of samples wrongly classified to this class is $S_F^i$. Then the MAP can be expressed as:

\begin{equation}
MAP = \frac{1}{k} \sum_{i=0}^{k-1} \frac{S_T^i}{S_T^i+S_F^i}
\end{equation}

As shown in Table 1, the mixed motion stream obtains the worst performance which demonstrates the significance of motion separation in group activity recognition. For the effectiveness of fusing global and local motion streams, both accuracy and MAP have an obvious increase on the activities for 3-point, free-throw and slam dunk after the fusion of global and local motion streams. This can be explained by the fact that global and local motion patterns in the above three activities are complementary.

\begin{table}
\small
\caption{Group activity recognition results (Accuracy/Mean average precision(MAP)) on NCAA dataset \cite{1}. The results are reported on single global motion stream, single local motion stream, single mixed motion stream and global and local patterns fusion two-stream framework.}
\newcommand{\tabincell}[2]{\begin{tabular}{@{}#1@{}}#2\end{tabular}}
\begin{center}
\begin{tabular}{|c|c|c|c|c|c|}
\hline
\diagbox{Activity}{ACC/MAP}{Method} & \tabincell{c}{Mixed \\motion\\ Stream} & \tabincell{c}{Global \\ motion \\ Stream} & \tabincell{c}{Local \\motion \\ Stream} & \tabincell{c}{Two-Stream \\ Framework}\\
\hline\hline
3-point & 0.728/0.716 & 0.709/0.701 & 0.695/0.707 & \textbf{0.777}/\textbf{0.733} \\
\hline
Free Throw & 0.776/0.825 & 0.776/0.929 & 0.821/0.932 & \textbf{0.881}/\textbf{0.937}\\
\hline
Layup & 0.580/0.639 & 0.548/0.581 & \textbf{0.626}/0.618 & 0.609/\textbf{0.695}\\
\hline
2-point & 0.508/0.575 & 0.467/0.550 & \textbf{0.563}/\textbf{0.674} & 0.541/0.626\\
\hline
Slam Dunk & 0.333/0.08 & 0.500/0.138 & 0.389/0.09 & \textbf{0.556}/\textbf{0.147}\\
\hline
Steal & 0.921/0.899 & \textbf{0.978}/0.911 & 0.940/0.910 & 0.962/\textbf{0.924}\\
\hline\hline
Mean & 0.641/0.623 & 0.663/0.635 & 0.672/0.656 & \textbf{0.721}/\textbf{0.677}\\
\hline
\end{tabular}
\end{center}
\end{table}

\begin{figure}[t]
\begin{center}
   \includegraphics[width=0.8\linewidth]{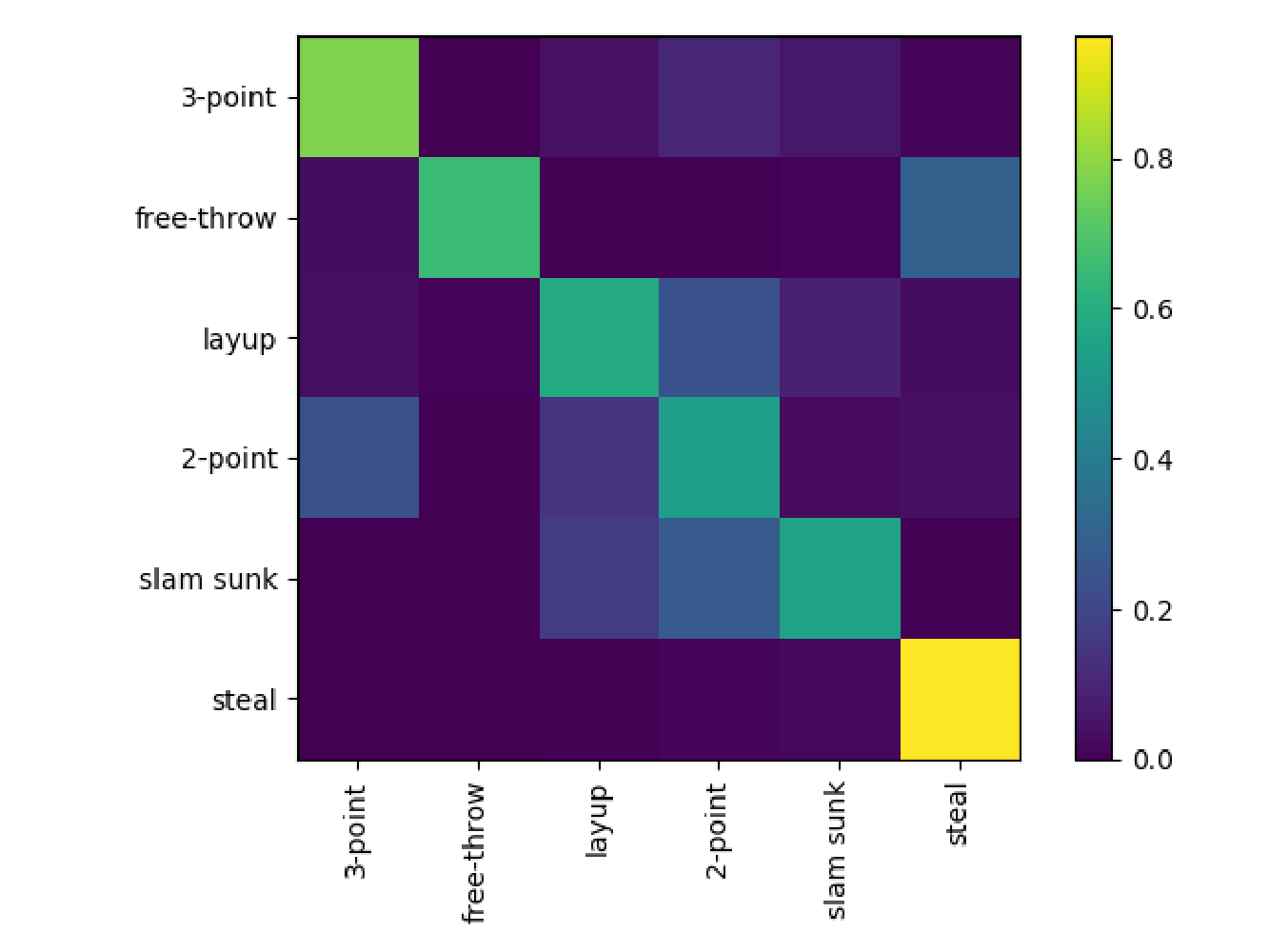}
\end{center}
   \caption{Confusion matrix for the NCAA dataset obtained using our two-stream framework.}
\end{figure}

Specifically, the global motion pattern in 3-point activity is similar with some of the 2-point activities. Nevertheless, local motion patterns can be more distinctive between 3-point and 2-point activities. The position distribution of players is relatively scattered in 3-point and rather denser in 2-point event. Therefore, local motion is more intensive in 2-point compared to that in 3-point. In this sense, the combination of these two modalities is beneficial for activity recognition.

The performance shows slightly reduction after the integration of two streams in layup and 2-point activities. In our analysis, these two activities share similar motion patterns in both global and local modalities. In this circumstances, fusion may not boost the performance. All models show excellent performance on steal event which is mainly because that motion direction reversal is an unique and discriminative feature in both global and local motion patterns among all the activities. Global motion is dominant in the scene thus the accuracy of global motion stream on steal activity achieves the best result. As expected, the mixed motion stream achieves similar results compared with global and local streams while is mainly because global motions are dominant in the scene. Thus the local motion patterns can not be fully generalized by the model. In addition, we draw the confusion matrix as shown in Figure 9 based on our global and local motion fusion model. Consequently, it can be demonstrated that learning feature representations separately on global and local motions is necessary and effective.

\subsection{Success/Failure Prediction}

Next, we evaluate our KVI based scheme for success/failure prediction. As mentioned in Section 5.2, given an image sequence, basket region is initially detected and the cropped regions are fed into CNN model to identify if the ball falls into the basket. The clip is predicted to be success as long as one of the frame in the clip is predicted to success. During the experiment, the frame is predicted to be the label of success when the response value of the corresponding neuron in the softmax layer is greater than a threshold. To obtain a proper value of the threshold, we adjust the threshold ranging from 0.5 to 1.0 with interval of 0.05. Under different thresholds, we test the accuracy on all success clips, all failure clips and overall testing clips. The accuracy curve shown in Figure 10 reveals that the highest performance of 0.887 is reached when the threshold is set to 0.7.

To demonstrate this view, we directly applied motion patterns to classify 11 classes of group activities with success/failure and the results are shown in Table 2. As shown in Table 2, the performance of using only motion patterns is much lower than using motion patterns + KVI.

\begin{figure}
\begin{center}
   \includegraphics[width=0.8\linewidth]{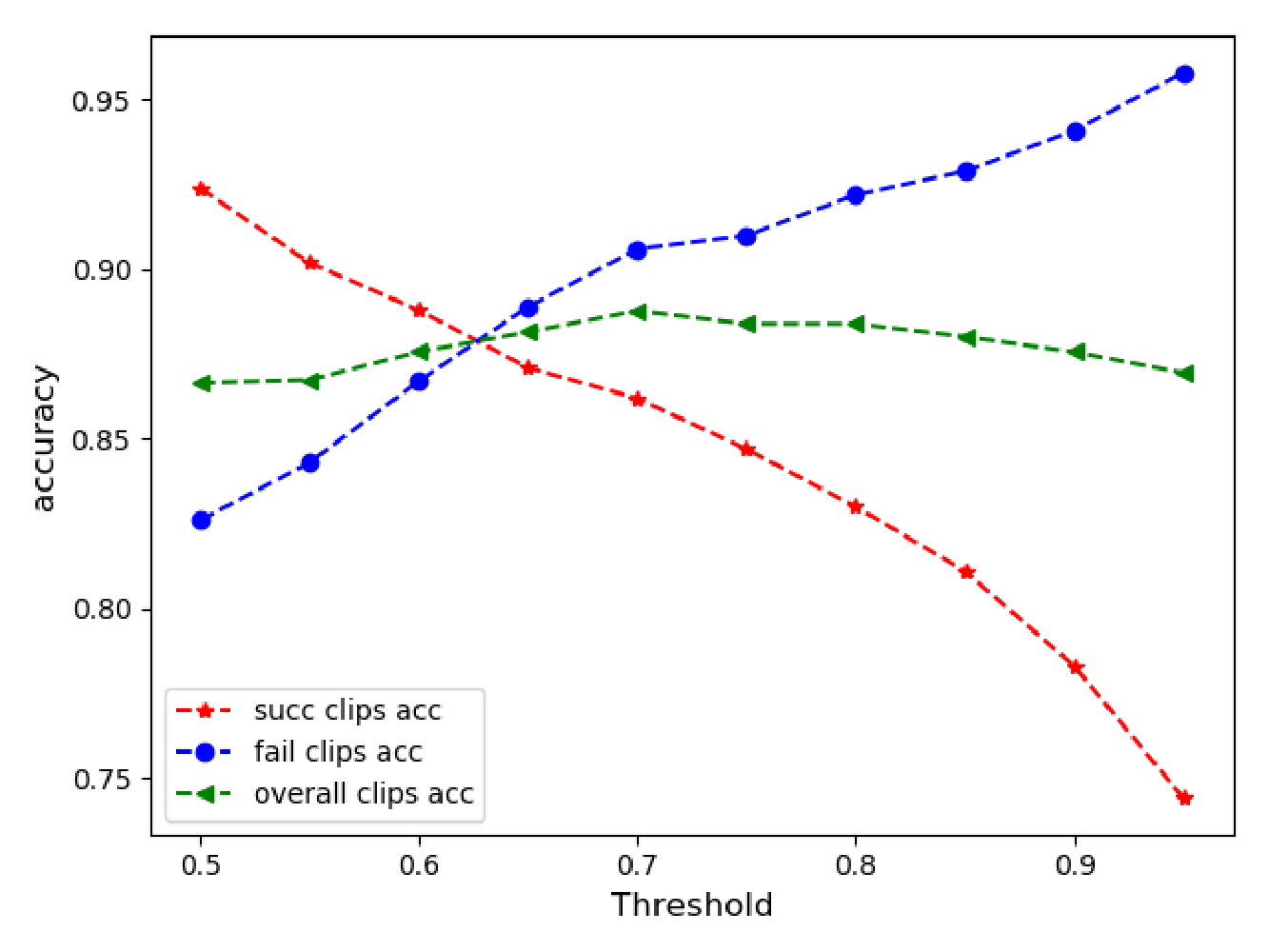}
\end{center}
   \caption{The accuracy curve of success clips, failure clips and overall activity success/failure prediction accuracy under different thresholds, on the NCAA dataset \cite{1}.}
\end{figure}

\begin{figure}
\centering
\subfigure[Success clip that is wrongly predicted to failure.]{
\label{Fig.sub.1}
\includegraphics[width=0.96\linewidth]{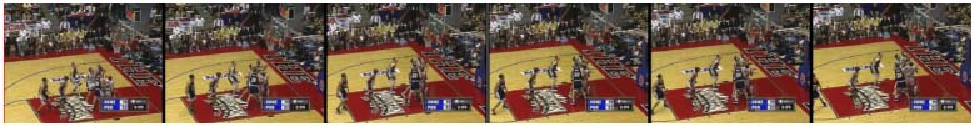}}
\subfigure[Failure clip that is wrongly predicted to success.]{
\label{Fig.sub.2}
\includegraphics[width=0.96\linewidth]{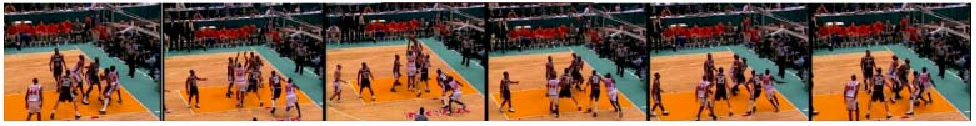}}
\caption{Examples of clips that are wrongly predicted. For simplicity, we only show the last 30\% of the frames which are highly correlated to shooting success and failure.}
\label{Fig.lable}
\end{figure}

\begin{table}
\scriptsize
\caption{Performance comparison on 11 classes semantic basketball events classification results based on method with/without key visual information (KVI).}
\newcommand{\tabincell}[2]{\begin{tabular}{@{}#1@{}}#2\end{tabular}}
\begin{center}
\begin{tabular}{|c|c|c|c|c|c|c|}
\hline
Method & \tabincell{c}{Global \\ Motion} & \tabincell{c}{Global Motion\\+ KVI} & \tabincell{c}{Local \\ Motion} & \tabincell{c}{Local Motion\\+ KVI} & Two-stream & \tabincell{c}{Two-stream \\ + KVI}\\
\hline
Accuracy & 0.381 & \textbf{0.628} & 0.403 & \textbf{0.657} & 0.409 & \textbf{0.712}\\
\hline
\end{tabular}
\end{center}
\end{table}

\begin{table}
\scriptsize
\caption{Performance (mean average precision) comparisons with the state-of-the-art methods on the NCAA dataset \cite{1} for basketball video semantic event recognition.}
\newcommand{\tabincell}[2]{\begin{tabular}{@{}#1@{}}#2\end{tabular}}
\begin{center}
\begin{tabular}{|c|c|c|c|c|c|c|c|c|c|}
\hline
\tabincell{c}{Semantic \\ Events} & \tabincell{c}{IDT \\ \cite{25}} & \tabincell{c}{C3D\\ \cite{33}} & \tabincell{c}{MIL\\ \cite{40}} & \tabincell{c}{LRCN \\ \cite{41}} & \tabincell{c}{BLSTM\\ \cite{1}} & \tabincell{c}{On\_GCMP \\ \cite{47}}  & \tabincell{c}{Global \\ + KVI}  & \tabincell{c}{Local \\+ KVI} & \tabincell{c}{Two-stream \\+ KVI}\\
\hline\hline
3-point succ. & 0.370 & 0.117 & 0.237 & 0.462 & 0.600 & 0.737 & 0.705 & 0.688 & \textbf{0.781} \\\hline
3-point fail. & 0.501 & 0.282 & 0.335 & 0.564 & 0.738 & 0.753 & 0.714 & 0.710 & \textbf{0.764} \\\hline
Free. succ. & 0.778 & 0.642 & 0.597 & 0.876 & 0.882 & 0.656 & 0.767 & 0.851 & \textbf{0.885} \\\hline
Free. fail. & 0.365 & 0.319 & 0.318 & 0.584 & 0.516 & 0.772 & 0.722 & 0.797 & \textbf{0.842} \\\hline
Layup succ. & 0.283 & 0.195 & 0.257 & 0.463 & 0.500 & 0.491 & 0.593 & 0.673 & \textbf{0.686} \\\hline
Layup fail. & 0.278 & 0.185 & 0.247 & 0.386 & 0.445 & \textbf{0.528} & 0.476 & 0.506 & 0.493 \\\hline
2-point succ. & 0.136 & 0.078 & 0.224 & 0.257 & 0.341 & \textbf{0.629} & 0.387 & 0.404 & 0.418 \\\hline
2-point fail. & 0.303 & 0.254 & 0.299 & 0.378 & 0.471 & \textbf{0.655} & 0.566 & 0.649 & 0.605 \\\hline
Slam. succ. & 0.197 & 0.047 & 0.112 & 0.285 & 0.291 & 0.308 & 0.591 & 0.681 & \textbf{0.724} \\\hline
Slam. fail. & 0.004 & 0.004 & 0.005 & 0.027 & 0.004 & \textbf{0.250} & 0.240 & 0.067 &0.194 \\\hline
Steal & 0.555 & 0.303 & 0.843 & 0.876 & 0.893 & 0.612 & \textbf{0.979} &0.941 & 0.963 \\\hline\hline
Mean & 0.343 & 0.365 & 0.221 & 0.316 & 0.516 & 0.581 & 0.613 & 0.630 & \textbf{0.669} \\
\hline
\end{tabular}
\end{center}
\end{table}

Samples that are wrongly predicted are shown in Figure 11. For the success clip that wrongly predicted to failure, the main reason is that the end point is annotated too early that the ball has not yet been in the basket. We believe that by extending the end point for short period of time, the performance can be further improved. For the failure clip that wrongly predicted to success, we think that the background noise and low resolution are the main reasons leading to misclassification. Also, in some cases, players may unintentionally hit the net and make the net move when blocking or struggling for rebound thus influence the results. With the augmentation of the training set by introducing detected basket images under various circumstances, this problem would be alleviated.

\subsection{Comparison with the State-of-the-Art}
We compare our proposed method against dense trajectories with Fisher encoding (IDT)~\cite{25}, C3D with SVM classifier (C3D) \cite{33}, multi-instance learning method (MIL)~\cite{40}, long-term recurrent convolutional networks (LRCN)~\cite{41}, attention based key player recognition method (BLSTM)~\cite{1} and On\_GCMP~\cite{47} as shown in Table 3. Our proposed MPs and KVI based methods outperform these baselines by a large margin of 15.3\%-44.8\% on average. We obtain better performance on all the events and show significant improvement on 3-point succ., Free-throw fail., layup succ. and slam dunk succ.. Overall, the superior performance indicates that our proposed scheme focuses on more discriminative features in basketball videos and is more robust and effective to background variations.

Then, we make a comparison with ontology embedded global and collective motion pattern based method (On\_GCMP)~\cite{47}. Similar to our perspective, \cite{47} used motion patterns as cues for semantic events modeling. However, they directly utilized mixed motion as input without motion separation which caused the loss of motion information. Although better performance is obtained on layup fail., 2-point succ., 2-point fail. and slam dunk succ. events. They extended the NCAA dataset forward and backward and employed a multi-stage framework that firstly merged 2-point and layup for 5-class prediction on event-occ stage, further classified 2-point and layup on pre-event stage, and finally judged success or failure on post-event stage. We only employ the data from NCAA dataset (event-occ) for both group activity recognition and success/failure prediction and outperform \cite{47} by 8.8\% on semantic event recognition.

\subsection{Generalization Ability of the Proposed Method}


\begin{table}
\scriptsize
\caption{Performance (Accuracy/Mean average precision(MAP)) comparison on NBA\&CBA dataset.}
\newcommand{\tabincell}[2]{\begin{tabular}{@{}#1@{}}#2\end{tabular}}
\begin{center}
\begin{tabular}{|c|c|c|c|c|c|}
\hline

Group Activity & GCMP & Mixed Motion & \tabincell{c}{Ours\\Global Motion} & \tabincell{c}{Ours\\Local Motion} & \tabincell{c}{Ours\\Two Stream}\\
\hline\hline
3-point & 0.600/0.564 & 0.747/0.612 & 0.726/0.590 & 0.779/0.536 & \textbf{0.863}/\textbf{0.651}\\
\hline
Free Throw & 0.214/1.000 & 0.428/1.000 & \textbf{0.500}/1.000 & 0.214/1.000 & \textbf{0.286}/1.000\\
\hline
Layup & 0.509/0.293 & 0.526/0.319 & 0.561/0.317 & 0.596/0.400 & \textbf{0.702}/0.400\\
\hline
2-point & 0.267/0.564 & 0.295/0.485 & 0.232/0.433 & 0.217/0.434 & \textbf{0.274}/\textbf{0.604}\\
\hline
Slam Dunk & 0/0 & 0/0 & 0/0 & 0/0 & 0/0\\
\hline
Steal & 0.909/0.233 & 0.909/1.000 & 0.909/1.000 & 0.909/1.000 & 0.909/1.000\\
\hline\hline
Mean & 0.417/0.442 & 0.484/0.569 & 0.488/0.557 & 0.453/0.557 & \textbf{0.506}/\textbf{0.609}\\
\hline
\end{tabular}
\end{center}
\end{table}

\begin{figure*}
\begin{center}
   \includegraphics[width=0.9\linewidth]{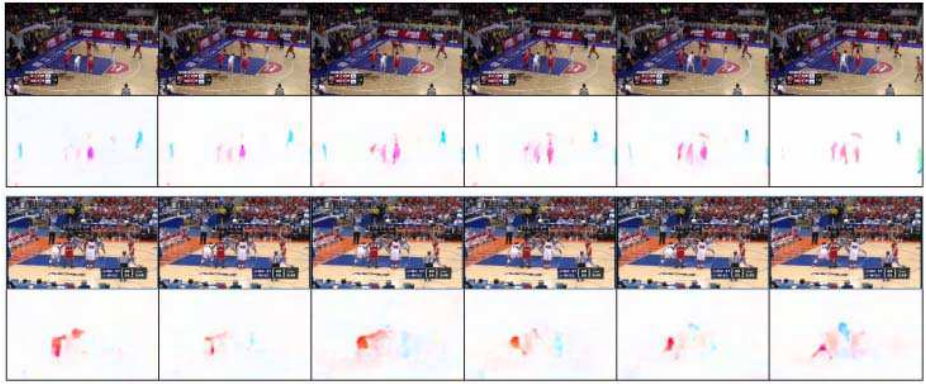}
\end{center}
   \caption{Visualization results of motions from free-throw activity. The first row is the video frames of free-throw from NBA\&CAB dataset and the second row is inter-frame motion fields. The third row is the video frames of free-throw from NCAA dataset and the last row is the corresponding inter-frame motion fields.}
\end{figure*}

\begin{figure}
\begin{center}
   \includegraphics[width=0.7\linewidth]{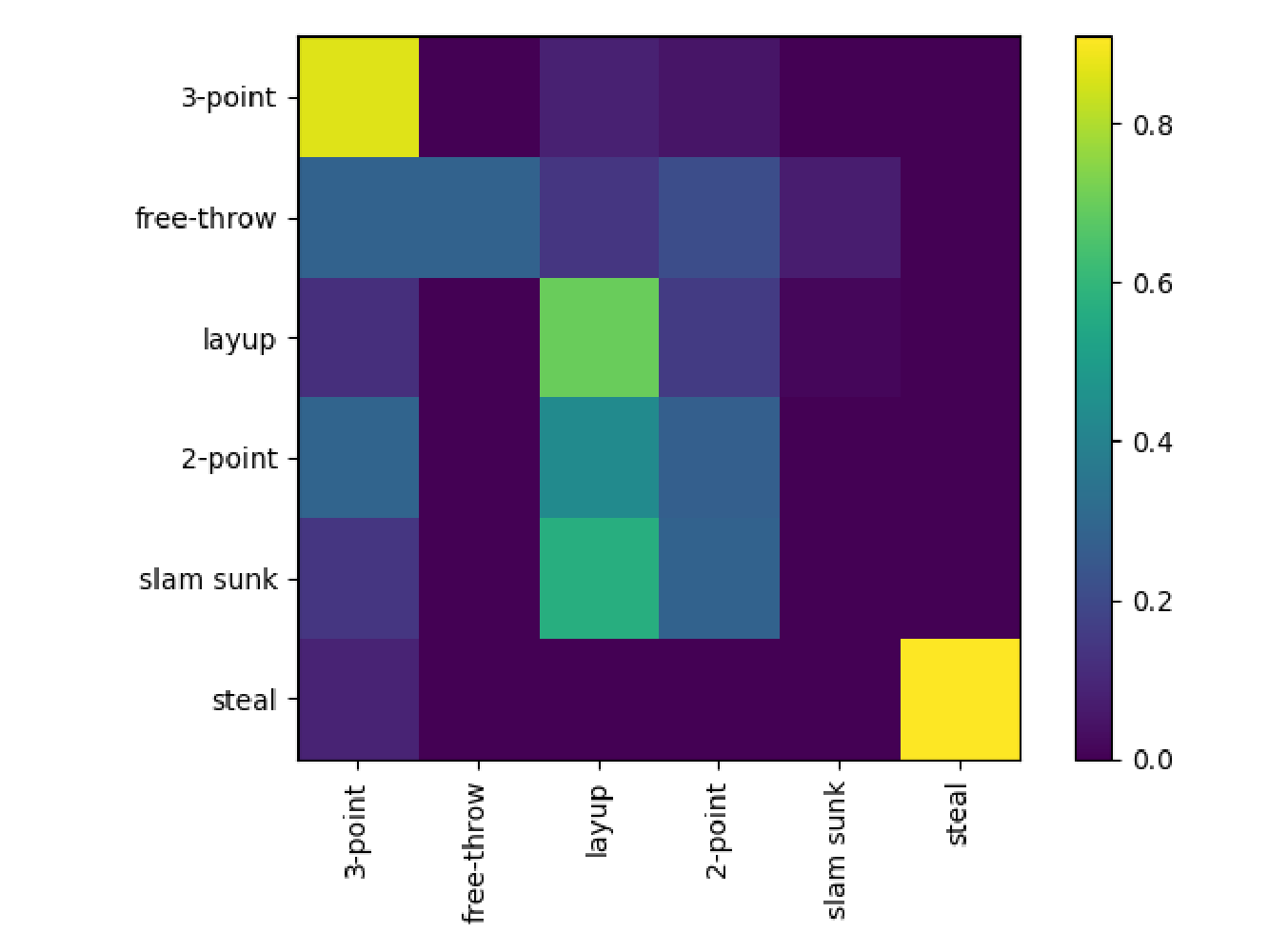}
\end{center}
   \caption{Confusion matrix for the NBA\&CBA dataset obtained using our two-stream framework.}
\end{figure}

In order to better verify the generalization ability of our proposed method, we further conduct experiments on NBA\&CBA dataset. We do not fine-tune on NBA\&CBA dataset and treat it as a testing set. The performance of our method reported in Table 4 is achieved by fusing the global and local stream network with weight of 1 : 1. For fair comparison, we reimplement the GCMP method \cite{47} with the NCAA training set. Our method achieves 50.6\% accuracy and 60.9\% MAP on NBA\&CBA dataset which outperform GCMP method by 8.9\% and 16.7\%.

In NBA\&CBA dataset, the camera position during free throw activity is farther away from the court that in NCAA dataset resulting in different representations of local motions as shown in Figure 12. However, the global motion is similar that the camera stays almost motionless during the free throw activity. It is difficult to predict free throw activity by mixed motion thus both our method and GCMP do not work well on this category. Our method obtains 28.6\% accuracy which is mainly owing to the contribution of the separated global motion stream. As shown in Figure 13, we can observe some failure cases in layup activity, which is probably due to the similar motion regulations among layup, 2-point and 3-point activities. The performance can be improved by fine-tuning on data from NBA or CBA basketball games. Both methods get poor performance on slam dunk activity due to the data imbalance problem that there is small amount of slam dunk clips in training set, failing to learning robust representations. From the experimental results, by separating global and local motion from the original motion, our scheme is able to explore more competitive feature representations from the dynamic scenes for group activity recognition.

\subsection{Runtime Analysis}

We test the runtime (ms) of each module on Titan X GPU as reported in Table 5. Motion estimation and motion separation modules work on parallel on frames in a video clip. Group Activity and success / failure classification module work on the batch of frames in a video clip. For each video clip, about 42 ms is needed which almost satisfies the real time application.

\begin{table}
\scriptsize
\caption{Runtime of each module in our framework.}
\newcommand{\tabincell}[2]{\begin{tabular}{@{}#1@{}}#2\end{tabular}}
\begin{center}
\begin{tabular}{|c|c|c|c|c|c|}
\hline

Module & \tabincell{c}{Motion\\Estimation} & \tabincell{c}{Motion\\Separation} & \tabincell{c}{Group Activity\\Classification} & \tabincell{c}{Succ./Fail. \\ Classification} & \tabincell{c}{Total}\\
\hline
\tabincell{c}{Runtime (ms)} & 7.14 & 11.2 & 6.3 & 17.2 & 41.8\\
\hline
\end{tabular}
\end{center}
\end{table}

\section{Conclusion}\label{sec:con}

We have proposed a novel approach to fuse the global and local motion pattern separation and key visual information for semantic event recognition in basketball videos. Through a two-pipeline framework, MPs and KVI are extracted for group activity recognition and for success/faillure prediction, respectively. Results from the two pipelines are further integrated using Kronecker product for semantic event recognition. Experimental results demonstrate that our proposed method is effective for semantic event recognition in basketball videos and obtain state-of-the-art performance on the NCAA dataset. The superior performance of cross dataset testing on NBA\&CBA dataset further demonstrates the generalization ability of our proposed method. In this work, we just separate the global and local motion, the local motion is still constrained by the camera movement. In future work, we will focus on completely removing the global motion and mapping the local motion to the real court.

\section{Acknowledgments}
This work was supported in part by the National Natural Science Foundation of China (61976010, 61802011, 61702022), Beijing Municipal Education Committee Science Foundation (KM201910005024), China Postdoctoral Science Foundation Funded Project (2018M640033), and "Ri Xin" Training Programme Foundation for the Talents by Beijing University of Technology.

\end{document}